\documentclass[12pt]{iopart}

\usepackage{color}
\usepackage{graphicx}
\usepackage{url}
\usepackage{hyperref}	
\usepackage{mathrsfs}
\usepackage{multirow}
\usepackage{cite}

\definecolor{greencolor}{rgb}{0,0.5,0.2}
\definecolor{redcolor}{rgb}{1.0,0.,0.}
\definecolor{bluecolor}{rgb}{0,0.,1.}
\definecolor{greycolor}{rgb}{.5,.5,.5}

\begin{document}

\title[]{Authorship recognition via fluctuation analysis of network topology and word intermittency}

\author{Diego R. Amancio}
\address{
Department of Computer Science \\
Institute of Mathematical and Computer Sciences\\
University of S\~{a}o Paulo, S\~{a}o Carlos, S\~ao Paulo, Brazil\\}
\ead{diego.raphael@gmail.com, diego@icmc.usp.br \\ }

\vspace{10pt}

\begin{abstract}
Statistical methods have been widely employed in many practical natural language processing applications. More specifically, complex networks concepts and methods from dynamical systems theory have been successfully applied to recognize stylistic patterns in written texts. Despite the large amount of studies devoted to represent texts with physical models, only a few studies have assessed the relevance of attributes derived from the analysis of stylistic fluctuations. Because fluctuations represent a pivotal factor for characterizing a myriad of real systems, this study focused on the analysis of the properties of stylistic fluctuations in texts via topological analysis of complex networks and intermittency measurements. The results showed that different authors display distinct fluctuation patterns. In particular, it was found that it is possible to identify the authorship of books using the intermittency of specific words. Taken together, the results described here suggest that the patterns found in stylistic fluctuations could be used to analyze other related complex systems. Furthermore, the discovery of novel patterns related to textual stylistic fluctuations indicates that these patterns could be useful to improve the  state of the art of many stylistic-based natural language processing tasks.
\end{abstract}

%
%
%
%
%

\section{Introduction}

The application of concepts from Physics in textual analysis has increasingly become widespread~\cite{Cristina,Ausloos,Martinez-Romo,Mehri,Isern,njp,huang}. The use of entropy concepts is perhaps one of the most known examples of adapting methods from Physics in language-based models~\cite{maxent}. In recent years, physicists have proposed novel approaches to tackle several natural language processing problems~\cite{sienkiewicz,wsenses,Yang,interplay,ausloos2,Yasseri,lcomplexity,voynich,consistency,wentian,xesus}. The emergence of fundamental principles of organization common to all languages has been studied in terms of the least-effort principle~\cite{canchopnas}. Other studies have been devoted to the analysis of word frequency distributions~\cite{Allahverdyan,Baixeries,Murtra,Eliazar,Elvevaa}, which has led to the design of novel cutting-edge keyword detection methods~\cite{voynich,Mathiesen,Carretero,Carpena,Herrera,Ortuno}.
Syntactical features have been employed to investigate the fundamental properties of the language from a physical standpoint~\cite{voynich,chc,wd}. In the semantic/pragmatic level, concepts from Physics have also been used to investigate the ubiquity of ambiguous structures in texts~\cite{wsenses,silva,semantic}.

In the field of stylometry, the use of complex networks (CN) in textual models has become commonplace~\cite{interplay,lcomplexity,voynich,comparing,Sheng,matching}. More specifically, several studies have modeled texts as co-occurrence (word adjacency) networks, where nodes and edges are represented by words and adjacency relationships, respectively. It has been shown that networks modeling texts share the same statistical properties of many other real systems~\cite{swhl}. Specially, such networks display both small-world and scale-free properties, as a consequence of the Zipf's law. Practical studies involving co-occurrence networks have devised algorithms to generate summaries~\cite{extractive}, to assess text coherence and cohesion~\cite{lantiq}, and to evaluate the quality of manual and machine translations~\cite{cnanalysis}. Even though word adjacency networks mostly grasp the syntactical factors of the language~\cite{voynich}, it has been shown that they also convey semantic information~\cite{wsenses,silva,semantic}.

While co-occurrence networks focus mainly on short scales, other physical models have been devised to capture long-range correlations. One of the most popular methods borrowed from the study of dynamical systems is the burstiness of word occurrences~\cite{Carretero}, which represents an attribute capable of capturing long-range textual features. Particularly, it has been shown that core words are unevenly distributed, while function words display distributions generated from random processes~\cite{Herrera}. Such findings have motivated the proposition of algorithms aiming to detect keywords in single texts~\cite{Carpena} using level statistics~\cite{comparing} and information theory~\cite{Carretero}. The long-range textual structure has also been studied at the character unigram level~\cite{localHistGram,tomovic}.

Most of the research on textual pattern recognition has focused on the search for recurring patterns in order to infer a specific class to unknown instances~\cite{Kudo}. This approach has certainly worked well as many enlightening findings have been made this way.
Despite the great number of studies on textual pattern recognition, the analysis of stylistic fluctuations along texts has received comparatively little attention.
Empirical studies of some real systems have shown that fluctuations play a pivotal role on the unambiguous characterization of complex systems~\cite{taylor,response,Kalyuzhny,Chen,Blumm}. For example, when topology is a relevant network feature, the most informative patterns might be hidden in outlier fluctuations~\cite{beyond}. If one considers the distribution of word frequency, the fluctuations around the average might be useful to detect the most relevant concepts~\cite{Carretero}. In biological systems, dynamical fluctuations of vital signals provide valuable information about the current state of the system~\cite{onvariation}.

Given the importance of the fluctuations in other real systems, the current paper presents a study on the properties of the stylistic variability along texts. Authors' styles were characterized upon measuring the topological connectivity of networks modeling texts~\cite{comparing}. The stylistic evolution was quantified upon splitting the texts in shorter subtexts, which in turn were represented as smaller networks. From the topological analysis of these varying networks, several interesting finding could be found. First and foremost, it was possible to identify the correct authorship of texts from a multivariate analysis of the stylistic fluctuations along literary works. Interestingly, in this model, the variability of the average shortest path lengths along subtexts turned out be the most relevant feature for discriminating distinct authors. To identify the authorship of books, the proposed model also took advantage of the intermittency of time series representing the spatial distribution of words.
Similarly to the CN-based model, a significant accuracy rate in discriminating authorship was found. Surprisingly, when the intermittency of $100$ functional words was employed as features of the classifiers, the precise authorship could be found in 65\% of the cases. As I shall show, the discovery of novel patterns related to the stylistic fluctuations in texts indicate that the proposed methodology can be extended to analyze other complex systems.

This paper is organized as follows. In Section \ref{m:methods}, the methods employed to represent texts as networks are presented. This section also swiftly presents the main topological measurements employed for the characterization of complex networks. In the same section, the intermittency concept is presented. In Section \ref{resultados}, the authorship recognition task is studied. In this case, the variability of complex network measurements along texts and the intermittency of specific function words were employed as attributes of the classifiers for the authorship recognition task. Finally, Section \ref{conclusao} presents perspectives for further research.

\section{Methods} \label{m:methods}

In this paper, the style of written texts was quantified by measuring the topological properties of complex networks~\cite{identification}. The representation of a text as a co-occurrence (word adjacency) network is detailed in Section \ref{netSec}. The topological features of complex networks employed to analyze the stylistic variation of texts are presented in Section \ref{medidas}. An alternative model based on the spatial distribution of words is presented in Section \ref{intDet}.

\subsection{Modeling texts as complex networks} \label{netSec}

There are several ways to model texts as networks~\cite{lnet}. While semantic networks capture the relationships between word meanings, co-occurrence networks are more suitable to grasp stylistic attributes of written texts. As a matter of fact, co-occurrence networks represent a simplified version of syntactic networks~\cite{voynich} because most of the syntactic connections occurs between neighboring words ~\cite{cancho,voynich}.

Prior to the creation of a co-occurrence network, some pre-processing steps are usually performed. Firstly, words conveying low semantic context (\emph{stopwords}) are removed. Most of the words considered as \emph{stopwords} are articles and prepositions {(see the Supplementary Information)}. They are removed from the analysis because such words are mostly employed to connect other content words. After removing the \emph{stopwords}, words with distinct spelling referring to the same concept are mapped to the same form. As a consequence, nouns and verbs are mapped to their singular and infinitive forms, respectively~\cite{manning}. To perform such mapping, it is imperative to solve ambiguities at the word level because the mapped form might depend upon the sense assumed for a given word in a given context. To assist the disambiguation algorithm, the words are labeled with their respective parts-of-speech~\cite{manning}. The labeling method employed is based on the maximum-entropy model proposed in \cite{kohavi}.

\begin{figure}[!htbp]
\begin{center}
    \includegraphics[width=0.4\linewidth]{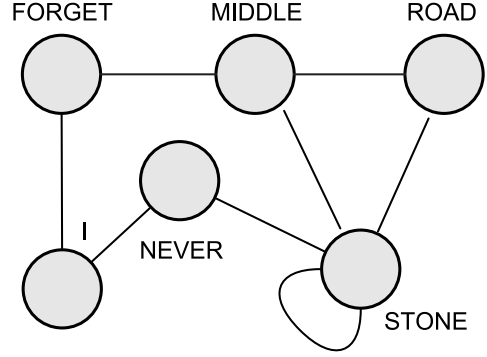}
        \caption{\label{f:redinhaexemplo}Example of co-occurrence network created for the poem ``In the middle of the road'', by Carlos Drummond de Andrade (see Table \ref{stoplem}). Note that, after the removal of \emph{stopwords}, adjacent words are connected (see first column of Table \ref{stoplem}).
    }
	\end{center}
\end{figure}

After the pre-processing step, each distinct word becomes a node. Therefore, the total number of nodes in the network is equal to the vocabulary size ($M$) of the pre-processed text. The words that appear separated by up to $d-1$ intermediate words are connected in the network.
In this paper, the value $d=1$ was used. Therefore, only  adjacent words were connected. Table \ref{stoplem} illustrates the pre-processing steps taken to form a small network from the poem ``In the middle of the road'', by Carlos Drummond de Andrade. The network obtained from the pre-processed form is shown in Figure \ref{f:redinhaexemplo}.
\begin{table}
	\centering
\caption{\label{stoplem}Example of pre-processing  steps performed to create a co-occurrence network. Firstly, \emph{stopwords} are removed (see step \#1). Then, the remaining words are converted to their canonical forms (see step \#2). As a consequence, nouns and verbs are mapped to their singular and infinitive forms, respectively.}
	\begin{tabular}{@{}lll}
            \hline
            {\bf Original text} & {\bf Step \#1} & {\bf Step \#2} \\
            \hline
            In the middle of the road 		& middle road       & middle road \\
            there was a stone there was 	& stone 			& stone \\
            a stone in the middle of 		& stone middle 		& stone middle \\
            the road there was a stone 		& road stone        & road stone \\
            in the middle of the road   	& middle road       & middle road \\
            there was a stone. Never 		& stone never       & stone never \\
            should I forget this event 		& I forget event    & I forget event \\
            in the life of my fatigued 		& life fatigued     & life fatigue \\
            retinas. Never should I 		& retinas never I   & retina never I \\
            forget that in the middle 		& forget middle     & forget middle \\
            of the road there was a 		& road              & road \\
            stone there was a stone  		& stone stone       & stone stone \\
            in the middle of the road 		& middle road       & middle road \\
            in the middle of the road 		& middle road       & middle road \\
            there was a stone.              & stone             & stone \\
			\hline
		\end{tabular}
\end{table}

\subsection{Topological characterization of complex networks} \label{medidas}

There are a myriad of measurements currently employed to characterize the topology of complex networks~\cite{newmanbook}. Traditional measurements can be classified according to the amount of information needed for the computation. While local measurements only require information about the neighbors of a given node, global measurements require that the global network connectivity is known beforehand. There is also a third class: the quasi-local measurements. As the name suggests, quasi-local measurements require information about further neighbors (i.e. the nodes located two or more hops away from the node under analysis).
The following list swiftly describes the measurements employed to analyze the topology of networks modeling texts.
\begin{itemize}

  \item \emph{Clustering coefficient}: the clustering coefficient ($C$), a quasi-local measurement, quantifies the density of links between the neighbors of a given node. If $c_i$ represents the number of edges between the neighbors of the node $v_i$ and $k_i$ is the total number of neighbors of $v_i$, the clustering coefficient is given by $C_i = 2 c_i  ( k_i^2 - k_i)^{-1}$. In co-occurrence networks, the clustering coefficient measures the number of contexts in which a given word appears~\cite{comparing}. While generic words tend to take low values of clustering coefficient, context-specific words usually take higher values of clustering coefficient~\cite{comparing}.

  \item \emph{Average shortest path length}: to define this global measurement, consider we are given $D_{ij}$, the shortest distance between nodes $v_i$ and $v_j$. The average shortest path length of $v_i$ is then given by
      \begin{equation}
        l_i =  \frac{1}{M(M-1)} \sum_i \sum_j D_{ij}.
      \end{equation}
      In textual networks, this measurement quantifies the relevance of words. More specifically, a given word is considered relevant either when it is highly frequent or when it occurs close to the most relevant words~\cite{comparing}.

  \item \emph{Betweenness}: the betweenness ($B$) is a global measurement that measures the relevance of words~\cite{newmanbook}. To do so, the betweenness quantifies the number of shortest paths passing through a specific node. In textual networks, the betweenness also quantifies the number of contexts in which a word appear~\cite{comparing}. However, unlike the clustering coefficient, this measurement uses the global network information to infer the specificity of a word.

  \item \emph{Accessibility}:  this measurement is an extension of the degree $k$~\cite{extractive}. To define the accessibility, consider that $p_{ij}^{(h)}$ is the probability of a random walker to go from node $v_i$ to node $v_j$ in $h$ steps. Mathematically, the accessibility ($\alpha$) is computed from the irregularity (entropy) of the distribution of $p^{(h)}$
      \begin{equation}
        \alpha_i^{(h)} = \exp \Bigg{(} - \sum p_{ij}^{(h)} \ln p_{ij}^{(h)} \Bigg{)}.
      \end{equation}
      In general terms, the accessibility has been proven useful to identify the borders of complex networks when self-avoiding random walks are performed~\cite{extractive}. In textual networks, this measurement has been employed to identify keywords and to generate informative extractive summaries~\cite{extractive}.

\end{itemize}

\subsection{Intermittency} \label{intDet}

In linguistic models, the effects of attraction and repulsion of words is an ever present phenomenon~\cite{17,Herrera}. Several studies have shown that the distribution of many words along documents is not regular~\cite{Carretero,Carpena,Herrera,Ortuno}. Particularly, keywords are usually unevenly distributed along texts~\cite{voynich,Mathiesen,Carretero,Carpena,Herrera,Ortuno}. This finding has motivated the design of keyword detection methods relying upon a single document~\cite{manning}. To analyze the spatial distribution of words, each token is mapped to a element in a temporal series.
The first word of the text represents the first element, the second word represents the second element and so forth. Given a word $w_i$ occurring $f_i$ times in the text, the recurrence times of $w_i$ generate the temporal series $T_i = \{t_1,t_2,t_3,\ldots,t_{f_i-1}\}$, where $t_1$ is the distance (i.e. the number of intermediary words) between the first and second occurrence of $w_i$, $t_2$ is the distance between the second and third occurrence of $w_i$ and so on.
Usually, two elements are added to the original temporal series $T_i$: the space $t_0$ until the first occurrence of $w_i$ and the space $t_{f_i}$ after the last occurrence of $w_i$. The distribution of $T_i$ might be characterized by the mean and standard deviation:
\begin{equation} \label{mean}
    \langle T \rangle_i = \frac{1}{f_i+1} \sum_{i=0}^{f_i} t_i = \frac{N+1}{f_i+1},
\end{equation}
\begin{equation} \label{deviation}
    \Delta T_i  =  \sqrt{ \frac{1}{n_i} \sum_{i=0}^{n_i} (t_i - \langle T \rangle) },
\end{equation}
where $N = \sum f_i$. Given $\langle T \rangle$ and $\Delta T$, the irregularity of the distribution $T_i$ is computed as
\begin{equation} \label{eq.int}
    I_i = \frac{\Delta T}{\langle T \rangle} = \frac{f_i+1}{N+1} \sqrt{ \frac{1}{f_i} \sum_{i=0}^{f_i} (t_i - \langle T \rangle) }.
\end{equation}
The measurement defined in eq. \ref{eq.int} is known as intermittency (or burstiness) of the distribution. It has been widely employed to detect keywords in texts as an alternative to the tf-idf technique~\cite{manning}. In addition, the intermittency has proven relevant to detect keywords in genetic sequences~\cite{Ortuno}.

\begin{figure*}[!htbp]
\begin{center}
    \includegraphics[width=0.85\linewidth]{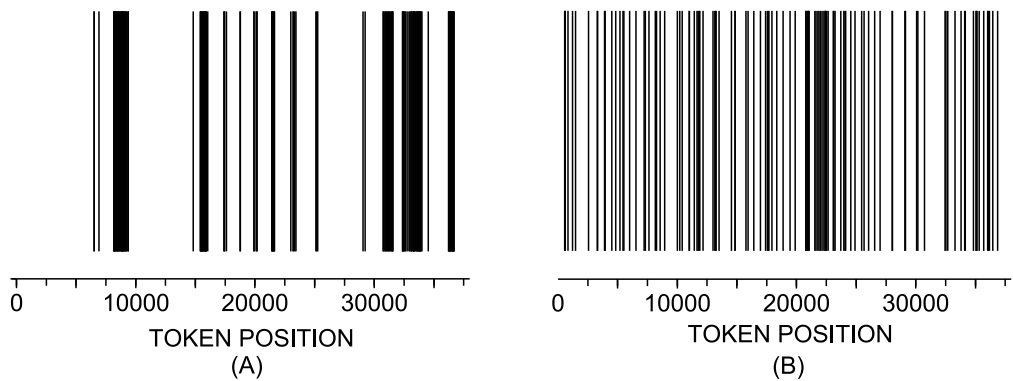}
        \caption{\label{f:barra}Profile of spatial distribution along the book ``Adventures of Sally'', by Pelham Grenville Wodehouse. The words considered were (a) ``\emph{Carmylle}''; and (b) ``\emph{feel}''. Note that the distribution of ``\emph{Carmylle}'' is much more irregular than the distribution of ``\emph{feel}''. This suggests that ``\emph{Carmylle}'' is much more relevant for the text than ``\emph{feel}''.}
	\end{center}
\end{figure*}

\begin{figure*}[!htbp]
\begin{center}
    \includegraphics[width=1\linewidth]{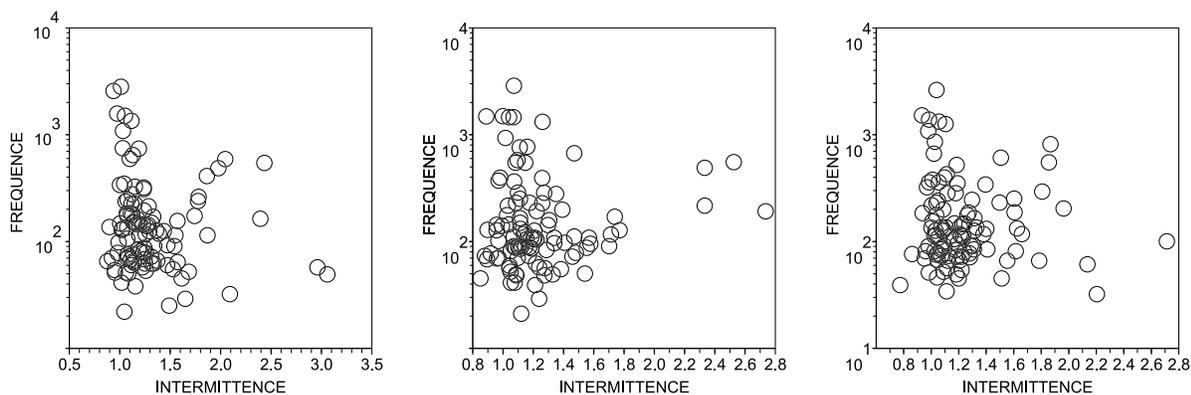}
        \caption{\label{f:cor}Pearson  correlation coefficient between intermittency and frequency in the books (a) ``Roughing It'', by  Mark Twain ($r=-0.14$); (b) ``The woman in white'', by  Wilkie Collins ($r=0.14$); and (c) ``Moby Dick'', by Herman Melville ($r=-0.17$).}
	\end{center}
\end{figure*}

A qualitative comparison of words taking distinct values of intermittency is provided in Figure \ref{f:barra}, which shows the distribution of the words ``\emph{Carmylle}'' ($f_i = 54$) and ``\emph{feel}'' ($f_i = 54$) along the book ``Adventures of Sally'', by Pelham Grenville Wodehouse. Because the distribution of ``\emph{Carmylle}'' is much more irregular than the distribution of ``\emph{feel}'', the former takes a much higher value of intermittency, as defined in eq. \ref{eq.int}. The burstiness revealed by ``\emph{Carmylle}'' also suggests that this word represents a relevant concept in the book~\cite{Herrera}. An important property of the intermittency measurement is that it does not correlate with the frequency (see Figure \ref{f:cor}). This means that the relevance assigned by the intermittency is not influenced by the word frequency. Taking advantage of this property, recent studies have combined both frequency and intermittency measurements to improve several keyword detection methods~\cite{voynich,Carpena}.

\subsection{Pattern recognition methods}

{The classification task aims at associating categories (or classes) to elements taking into account the attributes (or features) of these elements~\cite{nn}. More specifically, an attribute is a measurable property of objects.
To illustrate the concept, suppose that, in a given application, one desires to classify people according to their physical attributes. In this case, the height, the skin and hair color, the weight and others factors could be selected as attributes.
In many cases, the choice of discriminative and informative attributes plays an essential role on the performance of classification systems. Most of the attributes employed in traditional applications assume either numerical (e.g. $1$, $-7$ and $3.14$) or categorical values (e.g. \emph{low} and \emph{high}). In the current study, one of the attributes employed to characterize texts is the intermittency of specific words. In this case, the intermittency of each word represents a numeric attribute.
}

The classification task is of paramount relevance for information retrieval applications. Particularly, in this paper, pattern recognition methods are used to capture the patterns emerging from the representation of texts as networks. Moreover, pattern recognition methods are used to quantify the discriminative ability provided by these patterns. Currently, there are several automatic classification methods. They are traditionally divided into the following groups:
\begin{itemize}

  \item {\bf Supervised classification}: a binary relation mapping the input to the output is generated.

  \item {\bf Unsupervised classification}: a partition of the dataset is generated so that similar elements are clustered together.

  \item {\bf Semi-supervised classification}: the dataset available for automatic learning comprises a small set of labeled instances. Most of the instances is not labeled, i.e the class associated to these instances is lacking. In this case, the objective is to map the unlabeled input to a labeled output.

\end{itemize}

Typically, supervised classification methods process two datasets. The training dataset is the set of examples used as input. In other words, it represents the set of examples whose classes is known beforehand. In this paper, the training set is represented as $\mathcal{S}_{tr} = \{ \beta_{({tr},1)}, \beta_{({tr},2)}, \beta_{({tr},3)}\ldots\}$. The test dataset $\mathcal{S}_{ts} = \{ \beta_{({ts},1)}, \beta_{({ts},2)}, \beta_{({ts},3)}\ldots\}$ is the set used to evaluate the performance of the classifier.
A given example $\beta$  can be characterized by a set of $\mathcal{M}$ features: $\overrightarrow{\beta} = (F_1 = \beta^{(1)}, F_2 = \beta^{(2)},\ldots,F_\mathcal{M} = \beta^{(\mathcal{M})})$, where $\mathcal{F}=\{F_1,F_2,\ldots,F_\mathcal{M}\}$ is the set of attributes characterizing the example $\beta$. In other words, the $k$-th value taken by the attribute $F_k$ in $\beta$ is represented as $\beta^{(k)}$. In a supervised classification, a given example assumes a single class $c_i$ belonging to a finite set $\mathcal{C} = \{c_1,c_2,\ldots\}$.

{To quantify the quality of the classification, the cross validation technique was employed~\cite{weka}. In this method, a fraction of the dataset is used to perform the training and another fraction is used to perform the evaluation. The implementation of this technique consists in splitting the training dataset in ten folders. Initially, nine folders are selected to train the classifiers and the remaining one is used for evaluation. This process is repeated ten times so that a different folder is used for the evaluation in each iteration. Finally, the accuracy rate is computed as the average accuracy obtained over the ten iterations. The cross validation is considered a reliable index since the evaluation is performed over unknown instances.}

In the experiments, the analysis was performed on a dataset comprising books whose authorship is known beforehand. As a consequence, supervised classification methods were employed to recognize patterns in the generated textual time series. The methods employed in this study were: Bayesian Networks (BNT), Complement Naive Bayes (CNB), Naive Bayes (NVB), RBF Networks (RBF), Multi Layer Perceptron (MLP), Support Vector Machines (SVM), k Nearest Neighbors (KNN), C4.5 (C45) and Random Forest (RFO). A short introduction to these methods is provided in Appendix A.

\section{Results} \label{resultados}

The stylistic properties of texts was studied in the context of the authorship recognition task. In this problem, one tries to recognize the identity of authors whose authorship is unknown. Owing to its central importance for stylometry, several contributions have been proposed~\cite{asurveyon}. Simple approaches include the analysis of word length and additional character features~\cite{Tankard}. Mosteller and Wallace~\cite{Mosteller} proved that the frequency of function words (such as ``\emph{and}'', ``\emph{any}'', ``\emph{ever}'', ``\emph{or}'', ``\emph{until}'' and ``\emph{with}'') can be employed to quantify the style of authors.
More recently, many other approaches have been devised~\cite{asurveyon}, including those relying upon topological analysis of complex networks~\cite{comparing,someissues}. Here I use complex network and intermittency measurements to obtain potentially useful attributes for identifying the authors of books whose identity is lacking. Because this study focus on the analysis of stylistic fluctuations, the patterns displayed by the evolution of the statistical measurements along texts were studied.

Two types of temporal series representing the stylistic evolution in books are studied. In Section \ref{ZehVareia}, the relationship between the stylistic variation along books and the authorship recognition task is investigated. In Section \ref{viaInt}, the intermittency profile of some words across different authors is employed to perform the authorship recognition task. The dataset employed in the experiments comprises books written by 8 authors, as shown in {Table S1} of the Supplementary Information.

\subsection{Stylistic variation along books} \label{ZehVareia}

\begin{figure*}[!htbp]
\begin{center}
    \includegraphics[width=0.65\linewidth]{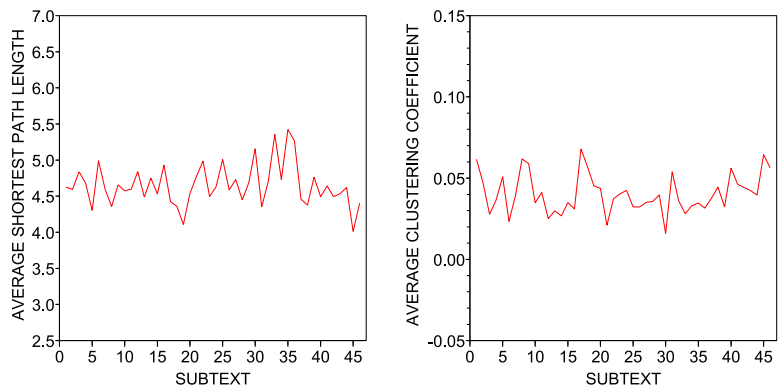}
        \caption{\label{f:var}Example of topological variation along the book ``Great Expectation'', by Charles Dickens. The networks were formed using $W=1,300$ tokens. The measurements considered were (a) the average shortest path length $\langle l \rangle$; and (b) the average clustering coefficient $\langle C \rangle$.
}
	\end{center}
\end{figure*}

In this section, I investigate whether the stylistic variation along texts provides useful attributes to the authorship recognition task. To quantify stylistic variations, the following methodology was taken. Each book in the dataset was split in subtexts comprising $W$ tokens. Assuming that a book is formed by a sequence of tokens $\mathcal{W} = \{w_1, w_2, \ldots\}$, the $j$-th subtext $\mathcal{T}_j$ will comprise the sequence $\{ w_{S_j}, w_{S_j+1}, \ldots, w_{S_j+W}\}$, where  $S_j = W \cdot j + 1$ and $j \in \{0,1,2,\ldots\}$. Each subtext
$\mathcal{T}_i$ was modeled as a complex network (see Section \ref{netSec}) and the topological measurements of each subtext were extracted (see Section \ref{medidas}). Thus, each topological measurement $X$ generates a temporal series $\mathcal{X} = \{x_1, x_2, \ldots x_P\}$, where $x_i$ represents the value obtained for $X$ in the subtext $\mathcal{T}_i$ and $P$ is the total number of subtexts.
An example of $\mathcal{X}$ for $X=\langle l \rangle$ and $X=\langle C \rangle$ is provided in Figure \ref{f:var}.
The temporal series $\mathcal{X}$ of each book was then decomposed in terms of the Fourier transform:
{
\begin{equation} \label{ft}
    \mathscr{F}(X)_{(j)} = \sum_{k=1}^{P}  x_k \exp \Bigg{(} \frac{-2\pi ijk}{N} \Bigg{)},
\end{equation}
}
where $i^2 = -1$. It is worth noting that the first component, given by
\begin{equation*}
    \mathscr{F}(X)_{(0)} = \sum_{k=1}^{P}  x_j = P \langle x \rangle,
\end{equation*}
only stores information concerning the average $\langle x \rangle$. Higher frequencies and, therefore, higher levels of variation in $\mathcal{X}$ are represented  in $\mathscr{F}(X)_{(\{ j \in \mathcal{N} | j\geq 1\})}$. As attributes of the classifiers, the first four components of $\mathscr{F}(X)$ were used.

The results obtained from the classification of authors are shown in Table \ref{tab.ar}.  The length of the subtexts considered were $W=\{500,~700,~900,~1,100,~1,300\}$. {For each subtext length, the table lists the accuracy rate obtained by the best classifier.} The lowest accuracy rate occurred for $W=500$ and the highest discriminability was achieved with $W=1,300$. In all cases, the performance obtained by the classifiers was statistically significant, as revealed by low $p$-values. This result confirms that the stylistic variations of authors along texts (quantified via topological analysis of complex networks) can be employed to discriminate authors' styles.
Specially, the proposed method could be used as a complementary stylistic attribute, because the stylistic variation has been widely neglected as a relevant feature in current authorship attribution methods~\cite{asurveyon}.
\begin{table}
\caption{Accuracy rate obtained in the classification based on the Fourier decomposition of time series of complex networks measurements. {For each subtext length ($W$), the table lists the accuracy rate obtained by the best classifier.}}
\label{tab.ar}
\begin{center}
\begin{tabular}{ccccl}
\hline
$W$ & {\bf Method} & {\bf Accuracy} & {\bf $p$-value} \\
\hline
{\bf 500} & BNT & 35.0\% & $2.2 \times 10^{-4}$  \\
{\bf 700} & CNB  & 37.5\% & $5.2 \times 10^{-5}$  \\
{\bf 900} & CNB  & 40.0\% & $1.1 \times 10^{-5}$  \\
{\bf 1,100} & RBF & 42.5\% & $2.2 \times 10^{-6}$  \\
{\bf 1,300} & RFO & 45.0\% & $4.0 \times 10^{-7}$  \\
\hline
\end{tabular}
\end{center}
\end{table}

To verify the relative relevance of the features employed in the authorship recognition task, the information gain of each attribute in the {training dataset} was computed. Mathematically, the relevance ascribed by the information gain ($\Omega$) is
\begin{equation} \label{infoganho}
    \Omega( \mathcal{S}_{{tr}}, F_k ) = \mathcal{H}(\mathcal{S}_{{tr}}) - \mathcal{H}(\mathcal{S}_{{tr}}|F_k),
   \end{equation}
where $\mathcal{H}(\mathcal{S}_{{tr}})$ is the entropy of the training dataset $\mathcal{S}_{{tr}}$ and $\mathcal{H}(\mathcal{S}_{{tr}}|F_k)$ is the entropy of $\mathcal{S}_{{tr}}$ when $F_k$ is specified. $\mathcal{H}(\mathcal{S}_{{tr}}|F_k)$ can be computed from training dataset as
{
\begin{equation}
     \mathcal{H}(\mathcal{S}_{tr}|F_k) =   \sum_{v \in V(F_k)}   { | \beta_{({tr})}^{(k)} = v | }  \cdot |\mathcal{S}_{{tr}} |^{-1} \cdot \mathcal{H}(  \beta_{({tr})}^{(k)} = v  ) ,
\end{equation}
}
where $| \cdot |$ is the cardinality of the set and $V(F_k)$ represents the set of all values taken by the attribute $F_k$ in the training dataset, i.e.
\begin{equation}
V(F_k) =  \bigcup_{i=1}^{ |\mathcal{S}_{{tr}}|} \beta_{({tr},i)}^{(k)}.
\end{equation}

\begin{figure*}[!htbp]
\begin{center}
    \includegraphics[width=0.75\linewidth]{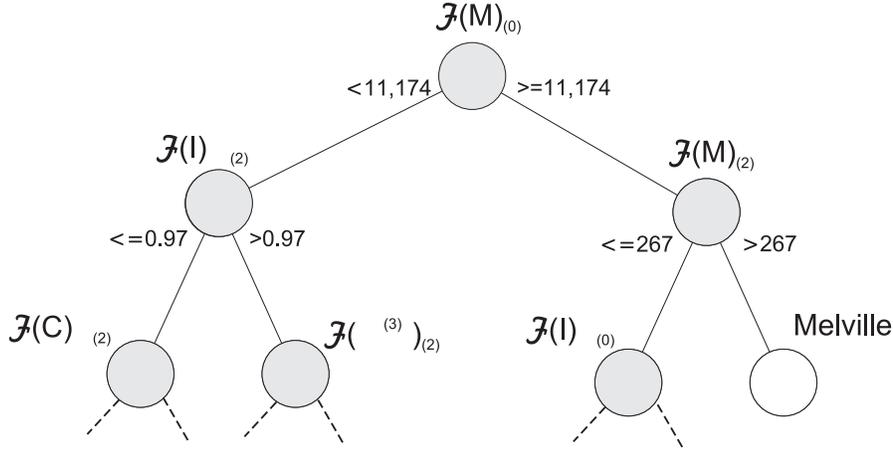}
        \caption{\label{f:treeExample}{Example of decision tree created to identify the authorship of books. To construct the tree, the C4.5 algorithm was employed in subtexts comprising $W=1,300$ tokens. Note that the second component of the average shortest path length and vocabulary size ($\mathscr{F}(\langle l \rangle)_{(2)}$ and $\mathscr{F}(\langle M \rangle)_{(2)}$)  are relevant as they appear at the top of the tree.}
    }
	\end{center}
\end{figure*}

The rank of the most informative measurements, according to eq. \ref{infoganho}, is shown in Table \ref{tab.relAtt}. {In this table, the rows indicate the ranking obtained by the attributes, for each subtext length ($W$).}
All in all, the vocabulary size $M$ turned out to be one of the most relevant attributes. The measurement displaying the highest relevance in higher components of the Fourier transform ($j\geq 1$ in eq. \ref{ft}) was the average shortest path length. More specifically, the third component ($j=3$) displayed the highest relevance for large values of $W$, suggesting that the attribute related to the variation of $\langle l \rangle$ along the text becomes even more relevant when larger subtexts are analyzed.
Interestingly, this result reinforces the importance of shortest paths for the authorship attribution task, since this measurement has been successfully employed to characterize authors' styles in networks formed from full books~\cite{comparing}.
The relevance of higher components of the Fourier transform can also be noted in the decision tree built with the C4.5 method~\cite{Quinlan} (see Figure \ref{f:treeExample}). Note that $\mathscr{F}(\langle l \rangle)_{(2)}$ and $\mathscr{F}(M)_{(2)}$ appear at superior levels of the tree, confirming thus their relevance.
The relative importance of higher components becomes even more apparent if one observes that some traditional complex network measurements (i.e.  $\mathscr{F}(X)_{(0)}$) correlates with the vocabulary size $M$~\cite{comparing}. {This does not occur with $\mathscr{F}(\langle l \rangle)_{(2)}$, as revealed by the Pearson correlation coefficient displayed in Table \ref{ascor}. In fact, none of the higher components found in Table~\ref{ascor} correlates significantly with other relevant traditional attributes ($\mathscr{F}(X)_{(0)}$), thus confirming that higher components indeed provide novel information for characterizing styles in written texts.}

The results concerning the evolution of styles revealed that distinct authors might display distinct stylistic patterns along texts. This finding is similar to the results found in~\cite{identification}, which showed that the temporal evolution of stylistic features of books published between $1590$ and $1922$ is able to identify the traditional literary movements. The main feature differentiating this work from previous studies is that the stylistic variation inside books is much more subtle than the corresponding variation over different literary styles~\cite{studyPlos}. The emergence of the described patterns suggests the applicability of other temporal models. Alternative models could probe, for example, the patterns present in the spatial distribution of character bigrams~\cite{localHistGram}. Particularly, this paper focus primarily on the evolution of stylistic patterns measured by the spatial distribution of words. For this reason, the next section investigates if the intermittency of specific words serves as authors' fingerprints for the authorship recognition task.
%
%

\begin{table*}
\caption{\label{tab.relAtt}Relative importance of the attributes used in the classification based on the spectral decomposition of complex network measurements. {The rows represent the ranking obtained for a given attribute. For example, the best attribute for $W=1,300$ was $\mathscr{F}(\langle C \rangle)_{(0)}$ and the second best attribute for $W=1,300$ was $\mathscr{F}(\langle M \rangle)_{(0)}$.}
The measurements taking values of information gain below $0.500$ are not shown. According to the information gain index, the third component of the average shortest paths lengths turned out to be one of the most informative measurement.
}
\centering
\begin{center}
\begin{tabular}{|c|c|c|c|c|c|}
\hline
\# & {\bf W=500} & {\bf W=700} & {\bf W=900} & {\bf W=1,100} &  {\bf W=1,300} \\
\hline
\multirow{2}{*}{1st} & $\mathscr{F}(M)_{(0)}$ & $\mathscr{F}(M)_{(0)}$ & $\mathscr{F}(\langle C \rangle)_{(0)}$ &  $\mathscr{F}(\langle l \rangle)_{(2)}$ & $\mathscr{F}(\langle C \rangle)_{(0)}$ \\
& 0.772 & 0.812 & 0.778 & 0.850 & 0.778 \\
\hline
\multirow{2}{*}{2nd} & $\mathscr{F}(\langle \alpha^{(3)} \rangle)_{(0)}$ & $\mathscr{F}(\langle C \rangle)_{(0)}$ 	& $\mathscr{F}(M)_{(0)}$ & $\mathscr{F}(M)_{(0)}$ & $\mathscr{F}(M)_{(0)}$ \\
& 0.669 & 0.778 & 0.772 & 0.772 & 0.772\\
\hline
\multirow{2}{*}{3rd} & $\mathscr{F}(\langle l \rangle)_{(0)}$   & $\mathscr{F}(\langle l \rangle)_{(0)}$  & $\mathscr{F}(\langle l \rangle)_{(0)}$	 & $\mathscr{F}(\langle l \rangle)_{(0)}$  &  $\mathscr{F}(\langle l \rangle)_{(0)}$ \\
& 0.665 & 0.772 & 0.712 & 0.691 & 0.712 \\
\hline
\multirow{2}{*}{4th} & $\mathscr{F}(\langle \alpha^{(2)} \rangle)_{(0)}$ & $\mathscr{F}(\langle \alpha^{(3)} \rangle)_{(0)}$ 	& $\mathscr{F}(\langle \alpha^{(3)} \rangle)_{(0)}$ & $\mathscr{F}(\langle \alpha^{(3)} \rangle)_{(0)}$ & $\mathscr{F}(\langle l \rangle)_{(2)}$\\
& 0.653 & 0.669 & 0.653 & 0.601 & 0.608 \\
\hline
\multirow{2}{*}{5th} & $\mathscr{F}(\langle l \rangle)_{(2)}$ & $\mathscr{F}(\langle \alpha^{(2)} \rangle)_{(0)}$ & $\mathscr{F}(\langle \alpha^{(2)} \rangle)_{(0)}$ & $\mathscr{F}(\langle \alpha^{(2)} \rangle)_{(0)}$ & $\mathscr{F}(\langle \alpha^{(3)} \rangle)_{(0)}$ \\
& 0.558  & 0.669 & 0.653 	& 0.601 &  0.606 \\
\hline
\multirow{2}{*}{6th} & & $\mathscr{F}(\langle l \rangle)_{(2)}$ & $\mathscr{F}(\langle l \rangle)_{(2)}$	& & $\mathscr{F}(\langle \alpha^{(2)} \rangle)_{(0)}$ \\
  & & 0.558 & 0.548 & & 0.601 \\
\hline
\multirow{2}{*}{7th} & & & & & $\mathscr{F}(\langle M \rangle)_{(2)}$ \\
  & & & & & 0.558 \\
\hline
\multirow{2}{*}{8th}  & & & & & $\mathscr{F}(\langle \alpha^{(3)} \rangle)_{(2)}$ \\																 &  & & & & 0.510 \\
\hline
\end{tabular}
\end{center}
\end{table*}

\begin{table}
\caption{\label{ascor}{
Pearson correlation coefficient $|r|$ between $\mathscr{F}(X)_{(j=0)}$ and the most informative measurements found for $W=1,300$ (see Table \ref{tab.relAtt}). Because all correlations assume low values, the information conveyed by $\mathscr{F}(X)_{(2)}$ differs from the simple average $\langle X \rangle = \mathscr{F}(X)_{(0)}$}.
}
\begin{center}
\begin{tabular}{llc}
\hline
$x$ & $y$ & $|r(x,y)|$ \\
\hline
$\mathscr{F}(\langle l \rangle)_{(2)}$ & $\mathscr{F}(\langle C \rangle)_{(0)}$ & 0.073  \\
$\mathscr{F}(\langle l \rangle)_{(2)}$ & $\mathscr{F}(M)_{(0)}$ & 0.182  \\
$\mathscr{F}(\langle l \rangle)_{(2)}$ & $\mathscr{F}(\langle l \rangle)_{(0)}$ & 0.170  \\
$\mathscr{F}(\langle l \rangle)_{(2)}$ & $\mathscr{F}(\langle \alpha^{(3)} \rangle)_{(0)}$ & 0.015  \\
$\mathscr{F}(\langle l \rangle)_{(2)}$ & $\mathscr{F}(\langle \alpha^{(2)} \rangle)_{(0)}$ & 0.100  \\
\hline
$\mathscr{F}(M)_{(2)}$ & $\mathscr{F}(\langle C \rangle)_{(0)}$ & 0.209  \\
$\mathscr{F}(M)_{(2)}$ & $\mathscr{F}(M)_{(0)}$ & 0.041  \\
$\mathscr{F}(M)_{(2)}$ & $\mathscr{F}(\langle l \rangle)_{(0)}$ & 0.036  \\
$\mathscr{F}(M)_{(2)}$ & $\mathscr{F}(\langle \alpha^{(3)} \rangle)_{(0)}$ & 0.059  \\
$\mathscr{F}(M)_{(2)}$ & $\mathscr{F}(\langle \alpha^{(2)} \rangle)_{(0)}$ & 0.115  \\
\hline
$\mathscr{F}(\langle \alpha^{(3)} \rangle)_{(2)}$ & $\mathscr{F}(\langle C \rangle)_{(0)}$ & 0.010  \\
$\mathscr{F}(\langle \alpha^{(3)} \rangle)_{(2)}$ & $\mathscr{F}(M)_{(0)}$ & 0.042  \\
$\mathscr{F}(\langle \alpha^{(3)} \rangle)_{(2)}$ & $\mathscr{F}(\langle l \rangle)_{(0)}$ & 0.045  \\
$\mathscr{F}(\langle \alpha^{(3)} \rangle)_{(2)}$ & $\mathscr{F}(\langle \alpha^{(3)} \rangle)_{(0)}$ & 0.080  \\
$\mathscr{F}(\langle \alpha^{(3)} \rangle)_{(2)}$ & $\mathscr{F}(\langle \alpha^{(2)} \rangle)_{(0)}$ & 0.077  \\
\hline
\end{tabular}
\end{center}
\end{table}

\subsection{Authorship recognition via word intermittency} \label{viaInt}

To verify if the uneven distribution of specific words along texts provides useful features for characterizing authors' styles, the following experiment was carried out. Following the research on stylometry, this study focused on function words. The 100 most frequent words in the corpus were considered as function words. As such, as attributes for the classifiers, the intermittency of these function words was used. The best classifier, the Multilayer Perceptron, yielded an accuracy rate of $65.0\%$ ($p$-value = $1.3 \times 10^{-14}$). This result suggests that, besides the frequency, the intermittency of specific function words might be useful for characterizing authors' styles in texts. Note that the discriminability obtained with intermittency features is not influenced by the frequency of function words, since there is no significant correlation between intermittency and frequency (see Section \ref{intDet}).

\begin{figure}[!htbp]
\begin{center}
    \includegraphics[width=0.65\linewidth]{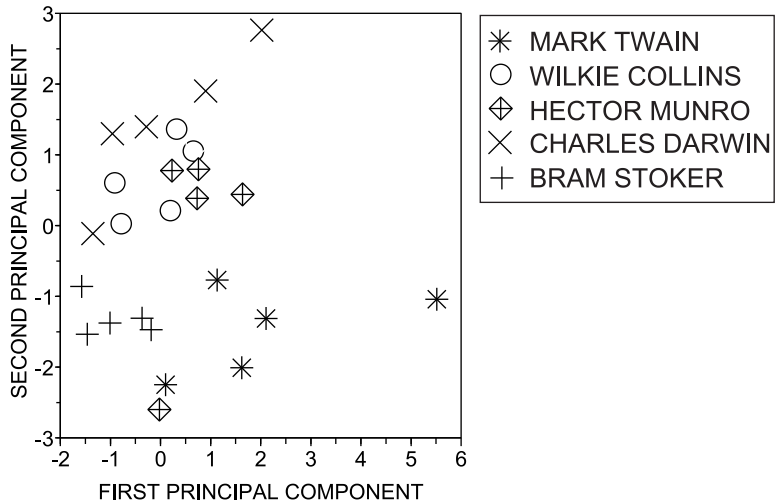}
        \caption{\label{f:pca_txi}Principal component analysis performed for the authorship recognition task. As attributes, only the five most informative features (the intermittency of ``\emph{but}'', ``\emph{and}'', ``\emph{I}'', ``\emph{who}'' and ``\emph{as}'') were employed to create the figure.
        %
        }
	\end{center}
\end{figure}

A detailed analysis of the classification revealed that most of the errors occurred for Arthur Conan Doyle, Wilkie Collins and Mark Twain (result not shown). If these authors are disregarded from the analyis, the use of intermittency features would provide an accuracy rate of 90\% with the Multilayer Perceptron.
Despite the large number of attributes employed for discriminating authors' styles, the discriminative ability concentrated in a few function words. According to the information gain measurement, the words displaying the highest discriminative ability were ``\emph{but}'' ($\mathcal{H} = 0.620$), ``\emph{and}'' ($\mathcal{H} = 0.604$), ``\emph{I}'' ($\mathcal{H} = 0.530$), ``\emph{who}'' ($\mathcal{H} = 0.494$) and ``\emph{as}'' ($\mathcal{H} = 0.462$). The high discriminability obtained with the intermittency of these five words can be noted in the principal component analysis shown in Figure \ref{f:pca_txi}.

{In summary, one can conclude that the representation of specific words as temporal series might be useful for the authorship recognition task. As commented in Section \ref{ZehVareia}, the use of intermittency of specific words combined with traditional features might be useful to improve the performance of style-based real applications.
In this case, an improved textual characterization would be provided, because the attributes generated from textual fluctuations do not correlate with traditional features. Moreover, distinct classifiers could be employed for each attribute type (e.g. frequency or intermittency), as some classifiers perform better for specific attributes. As such, the classification becomes more robust and accurate without the fine tuning required in single models~\cite{cocomin}. The combination of attributes could be performed via ordinary voting of simple models~\cite{bagging}. Another possibility is to consider fuzzy methods as independent classifiers and then select the best weighting strategy for each classifier~\cite{silva}.}
Furthermore, the successful application of intermittency measurements in characterizing authors' styles suggests that complementary studies should be carried out in order to probe whether additional features of temporal series modeling the spatial distribution of words are able to reveal novel stylistic/topological patterns.

\section{Conclusion} \label{conclusao}

In this study, I investigated if measurements characterizing temporal series from texts are useful to identify authors' styles.
In the light of the results, one can conclude that authors' stylistic properties can be characterized upon analyzing the fluctuations of textual statistical measurements. The statistically significant accuracy rates obtained in the authorship attribution task confirmed that the features derived from the fluctuation of specific {topological} and intermittency measurements are able to discriminate distinct authors. Using a co-occurrence network model, it was shown that the relative importance of distinct attributes may depend on the subtext length. Nevertheless, in general, further components of the Fourier decomposition of {topological measurements} turned out to be relevant features for the task. An analysis of the spatial distribution of specific words revealed distinct patterns of distribution for different authors. Surprisingly, the intermittence of functional words correctly discriminate the authorship in 65\% of the cases in a dataset comprising books written by 8 authors.

The focus of this investigation was on the evaluation of distinct attributes for characterizing authors' styles, rather than maximizing the accuracy rate of the classification. However, the dependence with stylistic attributes found for the proposed features suggests that attributes derived from the analysis of stylistic fluctuations can be combined in a hybrid way with traditional attributes, such as the frequency of function words~\cite{Mosteller}. As such, the findings reported in this paper shall potentially contribute to the improvement of current authorship recognition methods~\cite{Tankard}. One could pursue this line of analysis further, identifying the combination of features yielding the best discriminability.  Future investigations could probe the relevance of fluctuations in other related complex systems, such as DNA and other generic symbolic sequences, since the techniques described here can be extended in a straightforward fashion to such cases.

\section*{Acknowledgments}
DRA acknowledges financial support from S\~ao Paulo Research Foundation (FAPESP-Brazil) (grant number 2014/20830-0).

\newpage

\setcounter{section}{1}
\appendix

\section{Pattern Recognition Methods}

{This appendix swiftly describes the main pattern recognition methods employed in this study. A complete reference to the field of pattern recognition can be found in~\cite{bishop}}.

\subsection*{Decision trees}

\begin{figure}[!htbp]
\begin{center}
    \includegraphics[width=0.45\linewidth]{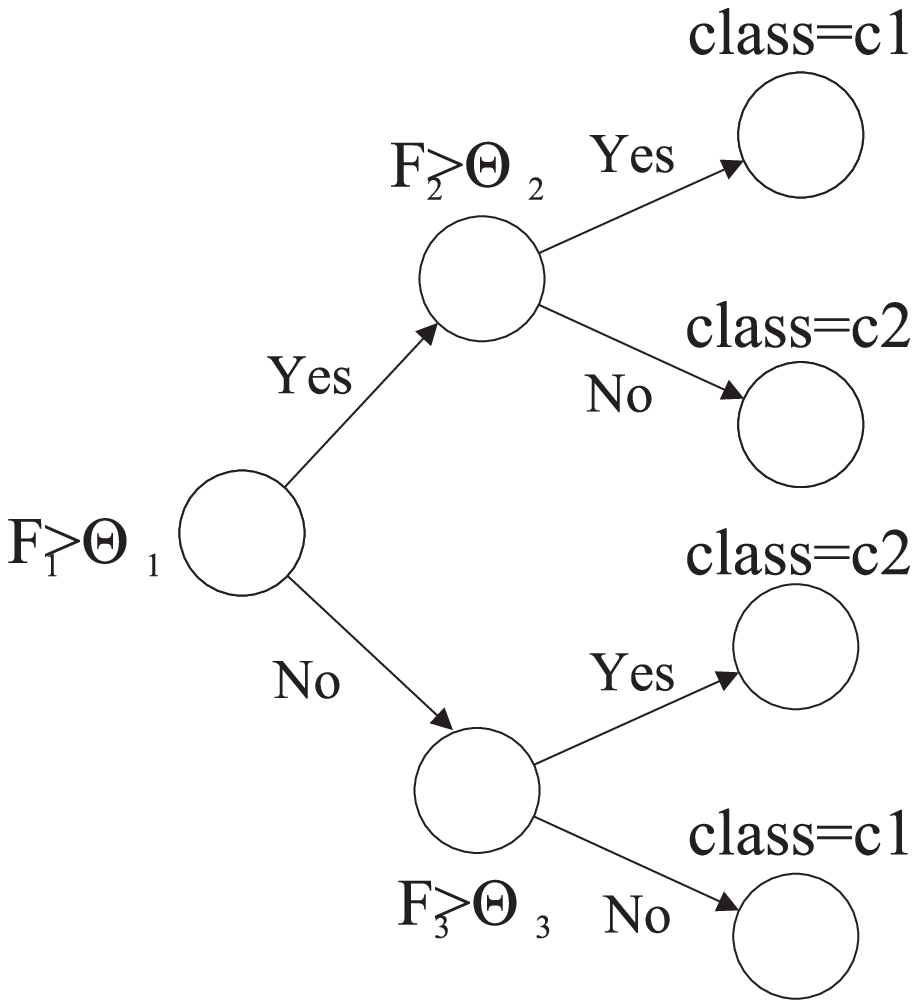}
        \caption{\label{f:treee}Example of decision tree. To classify a new instance, one starts the walk at the root node. The class assigned to the unknown instance is the class associated to the leaf node found at the end of the walk.
}
	\end{center}
\end{figure}

Decision tree algorithms employ trees~\cite{Cormen} to summarize the patterns recognized in the dataset (see Figure \ref{f:treee}). Typically, a decision tree comprises internal and leaf nodes. While internal nodes store the tests performed on specific attributes, leaf nodes represent classes. The edges connect nodes according to the answers obtained from the tests. For example, the node representing the test $F_1 > \theta_1$ has two outgoing edges, namely ``YES'' and ``NO''. During the classification stage, one travels through the tree until a leaf node is reached. In this case, the class associated to the leaf node is assigned to the unknown instance. The classification process is illustrated in Figure \ref{f:treee}.

To construct a decision tree, at each step, one tries to find an attribute $F_i$ and a threshold $\theta$ so that the test $F_i \geq \theta$ yields the best dataset partition. One assumes that the quality of a partition is proportional to the discriminability provided by that partition. At each division, the goal is to separate one or more classes in distinct groups. Several measurements have been proposed to quantify the quality of partitions. An well-known measurement is the Kullback-Leibler divergence~\cite{Kullback–Leibler}. The process of choosing the attribute with the highest information gain is reiterated for the two subsets created at each internal node. The recursion is finalized when a subset contains instances belonging to a single class. In this case, a leaf node is created to store the corresponding class.

The tree-based algorithms employed in this paper were the C4.5 and Random Forest. Further details regarding these methods can be found in~\cite{weka}.


\subsection*{Bayesian decision}

To classify a new instance, the Naive Bayes algorithm estimates the probability distribution of each class $c_i \in \mathcal{C}$. Given the likelihood profile of each class, the algorithm employs the \emph{maximum a posteriori} strategy to infer the correct class. The probability of each $c_i \in C$ to be assigned to the instance $\beta$ is
\begin{eqnarray*} \label{pprimeira}
    P(c_i | \overrightarrow{\beta} )  & = & \frac{P( \overrightarrow{\beta} | c_i) P(c_i)}{P(\overrightarrow{\beta})} \nonumber \\
    & = & \frac{ P( F_1 = \beta^{(1)}, \ldots, F_\mathcal{\mathcal{M}} = \beta^{(\mathcal{M})} | c_i ) P(c_i)}{P( F_1 = \beta^{(1)}, \ldots, F_\mathcal{M} = \beta^{(\mathcal{M})} )}.
\end{eqnarray*}
Note that $P(c_i)$ can be estimated as $\mathcal{N}(c_i) / \sum_{c_i \in \mathcal{C}} \mathcal{N}(c_i)$, where $\mathcal{N}(c_i)$
is the number of objects in $\mathcal{S}_{{tr}}$ belonging to class $c_i$.
For classification purposes, the quantity $P(\overrightarrow{\beta})$ can be disregarded from the analysis because $P( \overrightarrow{\beta} | c_i)$ is constant for all $c_i \in \mathcal{C}$. Finally, in order to estimate $P( \overrightarrow{\beta} | c_i)$, the traditional Naive Bayes classifier surmises independence between the features. Hence $P(\overrightarrow{\beta} | c_i)$ is estimated as
\begin{eqnarray*} 
 P( \overrightarrow{\beta} | c_i) & = & P( F_1 = \beta^{(1)}, \ldots, F_\mathcal{M} = \beta^{(\mathcal{M})} | c_i ) \nonumber \\
 & = & \prod_{k=1}^\mathcal{M} P(F_k = \beta^{(k)} | c_i).
\end{eqnarray*}
Using the value of $P( \overrightarrow{\beta} | c_i)$, it is possible to replace it in the definition of $P(c_i | \overrightarrow{\beta} )$. Therefore
\begin{equation*}
P(c_i | \overrightarrow{\beta} )  =   \frac{ P(c_i)}{ P( F_1 = \beta^{(1)}, \ldots)} \prod_{k = 1} ^ \mathcal{M} P(F_k = \beta^{(\mathcal{M})} | c_i).
\end{equation*}
Upon using the {\it maximum a posteriori} rule, the class $c_s$ can be estimated as
\begin{equation*} \label{phinal}
    c_{\beta} = \arg\max_{c_i \in \mathcal{C}} P(c_i) \prod_{k = 1}^\mathcal{M} P(F_k = \beta^{(k)} | c_i).
\end{equation*}
To obtain $c_{\beta}$ from the above equation, one must estimate the likelihood $P(F_k | c_i)$.  Several methods have been proposed to perform the estimation~\cite{estDist}. The Parzen-Rosenblatt window algorithm has been widely employed as a non-parametric technique to estimate probability densities~\cite{estDist}.

In addition to the Naive Bayes, the algorithms based on statistical paradigms employed in this study were the Complement Naive Bayes and Bayesian Networks. More details concerning these methods can be found in~\cite{weka}.

\subsection*{Neural Networks}

\begin{figure}
\begin{center}
    \includegraphics[width=0.32\linewidth]{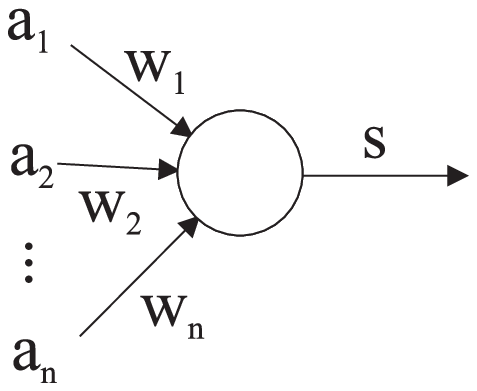}
        \caption{\label{f:neuronioZeh}Example of a single neuron. We are given some input signals $a_i$ and expected outputs in a supervised classification. The learning algorithm aims at minimizing the error between the actual and expected output signals.}
	\end{center}
\end{figure}

The simplest artificial neural network (ANN) model is the Perceptron. In this model, each neuron stores activation and transfer functions. While the former sums (with weights) the input signals, the latter yields an output signal as a function of the input.  Figure \ref{f:neuronioZeh} illustrates a single neuron with input signals and weights represented as $a_i$ and $w_i$, respectively. The output $s$ is $s = \sum_i a_i w_i + b$.
The transfer function $\phi$ may assume many distinct forms~\cite{nn}. A very simple possibility is to consider that the neuron is activated whenever $s$ surpasses a given threshold, i.e.
{
\begin{equation}
    \phi(s) = \left\{
    \begin{array}{c l}
        1 & \textrm{for } s > 0,\\
        0 & \textrm{otherwise.}
    \end{array}\right.
\end{equation}
}

The correct choice of synaptic weights in neural networks allows the network to effectively process the input signals in order to generate the expected output. In general, the weights are assigned by learning algorithms~\cite{nn}. Initially, the values $w_{ij}$ of weights linking the $i$-th node of the input layer with the $j$-th node of the output layer assume random values. Given these initial weights, several input signals are presented to the neuron. Then, the obtained output is compared with the expected values. If the observed error exceeds a given threshold, the current weights are modified by the learning algorithm. In this case, the larger the error obtained, the greater is the change applied to the current weights. More specifically, weights are updated according to the rule $w_{ij}^{(t+1)} = w_i^{(t)} + \eta \varepsilon_j x_i$,
%
%
%
where $\eta$ is the learning rate and  $\varepsilon_j$ is the error obtained for the $j$-th neuron.

The ANN-based pattern recognition methods employed in this study were the Multilayer Perceptron and the RBF network. More details concerning these methods can be found in~\cite{nn}.

\end{document}